# A decision support system for ship identification based on the curvature scale space representation

Álvaro Enríquez de Luna[1], Carlos Miravet[2], Deitze Otaduy[1], Carlos Dorronsoro[1]

[1]Centro de Investigación y Desarrollo de la Armada, Madrid, Spain
[2]SENER Ingeniería y Sistemas, S.A., Tres Cantos, Madrid, Spain

## ABSTRACT

In this paper, a decision support system for ship identification is presented. The system receives as input a silhouette of the vessel to be identified, previously extracted from a side view of the object. This view could have been acquired with imaging sensors operating at different spectral ranges (CCD, FLIR, image intensifier). The input silhouette is preprocessed and compared to those stored in a database, retrieving a small number of potential matches ranked by their similarity to the target silhouette. This set of potential matches is presented to the system operator, who makes the final ship identification. This system makes use of an evolved version of the Curvature Scale Space (CSS) representation. In the proposed approach, it is curvature extrema, instead of zero crossings, that are tracked during silhouette evolution, hence improving robustness and enabling to cope successfully with cases where the standard CCS representation is found to be unstable. Also, the use of local curvature was replaced with the more robust concept of lobe concavity, with significant additional gains in performance. Experimental results on actual operational imagery prove the excellent performance and robustness of the developed method.

## 1. INTRODUCTION

During the last years, there has been a sustained increase in the deployment of imaging sensors for surveillance and intelligence operations in naval scenarios, with sensors installed either on ground or aboard naval platforms. This increase in the operational sensor base highlights the need to reduce to a minimum the load of visual tasks performed by the system operator, in order to decrease the cost and to increase the reliability and the availability of the system.

A visual task of obvious importance is that of ship identification, which appears almost ubiquitously in naval surveillance operations. For years, this identification operation has been carried out by trained personnel, visually comparing the acquired silhouette to the reference ones stored in a database. All-visual identification is an error-prone operation requiring a painstaking effort, with operational and budgetary implications that could seriously hamper the increase of observation units at the pace required by current needs. In this context, it becomes apparent the need of a system with a high degree of automation that could assist the operator during the identification process, reducing the time to complete the task while simultaneously improving the reliability of the system.

In this paper, we describe the characteristics of a developed computer-based ship identification assistant developed by us, and report the performance figures obtained on actual operational imagery. The input data for the system is a silhouette of the vessel to be identified, which has been acquired from an approximate side view of the object. This silhouette is preprocessed and compared to those stored in a database, retrieving a small number of potential matches ranked by their similarity to the target silhouette. This set of potential matches is presented to the system operator, who then makes the final ship identification.

The system must cope with the large number of classes (> 1000, taking only military vessels into account) involved in this problem, and must be capable of operating on the imagery provided by different types of sensors (CCD, FLIR, image intensifier), and acquired under changing illumination and atmospheric conditions, at variable observation ranges. Moreover, the system must be tolerant regarding small variations in the observation angle, which lead to silhouettes that do not correspond to a perfect side view of the object. Finally, the system is required to produce its result in a limited amount of time (established in around 1 minute, for execution on a commercial PC platform) to ensure operational capacity.

To solve the abovementioned problem while meeting the necessary requirements, a system based on the use of the curvature scale-space (CSS) has been devised. The CSS transform[1, 2], which is part of the MPEG-7 standard, provides a robust mean to describe closed contours, and has been successfully used in several recognition tasks [3, 4], showing the ability to discriminate between a large number of classes, while simultaneously dismissing slight variations in shape that are perceptually irrelevant but unavoidable in any operational scenario.

The CSS representation describes the evolution in the location of the zero crossings of the silhouette curvature at different scales, which are obtained by convolving the silhouette with Gaussians of different variances. The smoothing operation induced by Gaussian convolution reduces both the sizes and the depths of the silhouette concavities and convexities. This reduction in size makes pairs of neighboring zero crossings to approach, ending up by collapsing when a certain degree of Gaussian smoothing is applied. Representing the evolution of the silhouette zero crossing locations with varying Gaussian smoothing yields a lobed figure, which is known as CSS image. For a thorough mathematical treatment of the CSS representation, refer to [2, 3].

The (x, y) coordinates of each lobe maximum in the CCS image represent, respectively, the location at the silhouette and the Gaussian kernel variance for which a collapse in neighboring zero crossings has occurred, and constitute useful features for classification in the standard CCS approach. Despite the power and robustness of this method, it entails certain characteristics that deem it as inadequate for its direct application to the problem to be solved. This has leaded us to incorporate certain modifications into the standard CCS method, which, in our environment, have lead to a significant increase in performance. Most notably, the CCS representation has been found to be unstable for a substantial number of vessel shapes. In these cases, slight variations in shape between the target and the model silhouettes have been found to induce abrupt changes such as lobe splitting in the corresponding CCS representations, thus precluding its use for reliable classification. To overcome this problem, the direct use of curvature zero crossings was abandoned in favor of curvature extrema (maximum/minimum), which have been found to be significantly more robust. Also, the use of local curvature was replaced with the more robust concept of lobe concavity, leading to a significant gain in performance. These modifications to the standard approach, together with some other additional changes are detailed in section 2.

Section 3 provides an outline of the complete algorithm, whereas section 4 shows the classification results on operational imagery, which prove the high performance and robustness of the developed method. Finally, some conclusions are drawn in section 5.

## 2. DRAWBACKS OF THE STANDARD CCS REPRESENTATION AND PROPOSED MODIFICATIONS. THE CCSS REPRESENTATION

In this section, we describe a number of problems that preclude the direct application of the CCS representation to our problem, together with the proposed solutions. This has lead to the implementation of a modified representation, which will be referred to in this paper as the Concavity-Convexity Scale Space (CCSS) representation. Major modifications are the tracking of curvature extrema instead of zero crossings, the choice of the lobe concavity concept instead of local curvature, the application of concavity-convexity thresholding to remove the effect of shallow concavities and the use of deck projection instead of arc length to describe feature location.

### 2.1. Instability of the CSS representation: the lobe splitting problem. Enhanced robustness by use of curvature extrema.

A basic requirement for silhouette representation is that it has to remain stable under slight variations in shape, as those commonly found between the silhouette of a target and that of its associated model. However, it has been found that the above requirement is not fulfilled for the CSS representation of a substantial number of vessel silhouettes. In those cases, small variations in shape, which are perceptually not relevant, can induce significant differences in the evolution of zero crossings, which end up by merging with different neighbors in target and model silhouettes. This yields CSS representations which are widely different in shape, despite the remarkable similarity of target and model in visual terms.

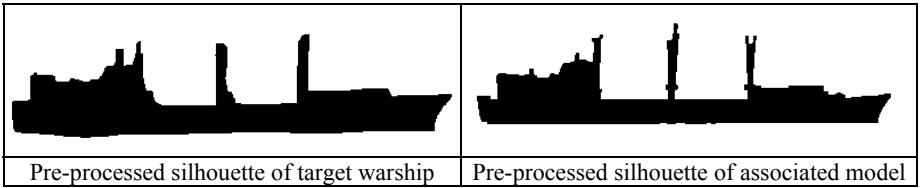
Fig. 1 Image and model silhouettes with lobe splitting problem

In Fig. 1, a typical example is presented, where the visual similarity between the preprocessed silhouettes of the warship image and the corresponding model can be appreciated. In principle, it would be expected that these similar shapes would result in CCS representations having a high degree of resemblance. The resulting CCS images are showed in Fig. 2, where a notorious difference in the disposition of lobes A and B becomes evident.

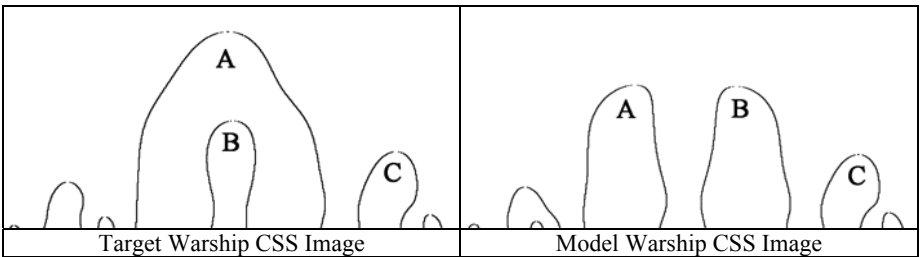
Fig. 2 Target and model CSS Images with lobe splitting problem

The cause of this divergent behavior can be easily traced back by examining the evolution of the zero crossing locations in both shapes. Fig. 3 shows the target and model silhouettes at three representative stages of the evolution, corresponding to convolution with Gaussians of increasing variance $\sigma_A$, $\sigma_B$ and $\sigma_C$. The location of the curvature zero crossings responsible for the generation of lobes A, B and C are also marked in each drawing.

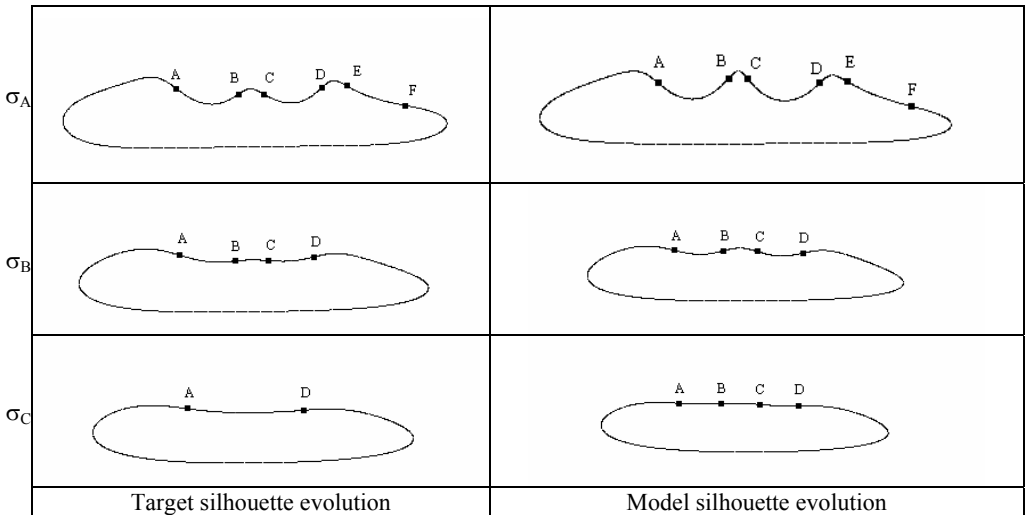
Fig. 3 Target and model silhouettes at three representative stages of their evolution

For the first selected stage in the evolution (variance $\sigma_A$), six curvature zero crossings (labeled from A to F) are found in the deck region of both silhouettes, delimiting a sequence of three concavities interleaved with two convexities. The visual appearance of both shapes is remarkably similar. At the next stage of the evolution, (variance $\sigma_B$), the E- and F- zero crossings have joined together in both the target and the model silhouettes, giving rise to lobe C of Fig. 2 for both the target and the model CSS image. The appearance of both shapes at this point in evolution is structurally identical, although with somewhat less prominent concavities and convexities in the target silhouette. Further evolution in the target silhouette causes zero crossings B and C to collapse, giving raise to lobe B in the target CSS image. Finally, zero crossings A and D merge, forming lobe A in the target CSS image. The evolution of the model silhouette at large smoothing factors is exactly the opposite, causing the collapse, respectively, of zero crossings A and B, and of C and D, leading to the formation of two separate lobes.

A considerable proportion of the training cases under analysis were affected at various degrees by this problem, pinpointing the need to seek a more stable silhouette representation. In this context, it is important to note that curvature extrema are known to be more stable than curvature inflection points (zero crossings) [5, 6], as they explicitly mark concavities or convexities in the shape. This sequence of concavities and convexities is generally preserved between a target and its corresponding model silhouette, even in cases where the lobe splitting problem appears in the CSS representation, as can be observed in Fig. 2 and Fig. 3. Hence, tracking of curvature extrema, instead of zero crossings, is proposed as a means to improve silhouette representation. This alternative representation also provides additional information concerning the curvature sign between neighboring zero crossings, which was not explicitly taken into account in the standard CSS representation, and which can be used to advantage in the identification task. Fig. 4 shows the improvement in stability obtained with the new representation, based on curvature extrema.

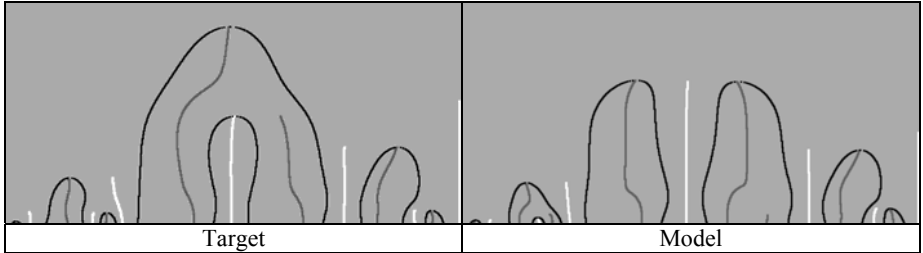

Fig. 4 CSS representation (black), and extrema CSS representation: maximum (white) and minimum CSS (dark gray)

**2.2. Model-to-target variations of the silhouette's arc length. Enhanced stability by deck projection**

The arc length of the ship's closed silhouette is normalized before comparison in both target and model. As the bottom part of the silhouette is usually featureless, the relative weight of the deck arc length will increase according to the level of detail of the deck features. Due to this fact, a feature on a certain deck position will appear at different locations (it will be shifted) in the maximum and minimum CSS images depending on the level of detail contained in the silhouette. In Fig. 5 this fact can be appreciated: the target-extracted silhouette is more detailed than the model one. In Fig. 6, panels a-b, the CSS images are showed for both cases, clearly revealing a shift in the corresponding curves.

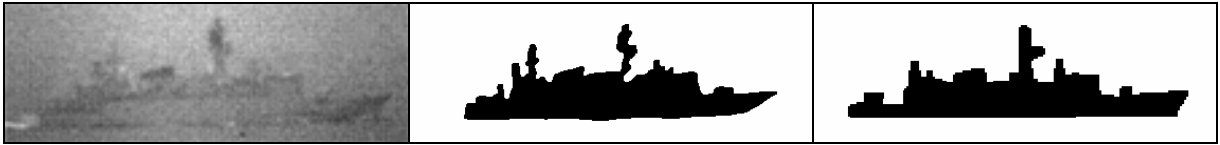

Fig. 5 Original target image, associated extracted silhouette and pre-processed model image.

Fig. 7 displays the target and model silhouettes at an intermediate stage of the evolution, showing the differences in the normalized arc length location of the main antenna in the deck. A solution to this problem can be simply obtained by projecting the locations of the curvature extrema onto horizontal coordinates of the deck, which are independent of arc length and, consequently, independent of the silhouette level of detail. In this sense, curvature extrema should be easier to compare and the results should be more accurate. Target and model extrema CSS images with deck projection are displayed in Fig. 6, panels c-d, showing a significant reduction in the problematic shift.

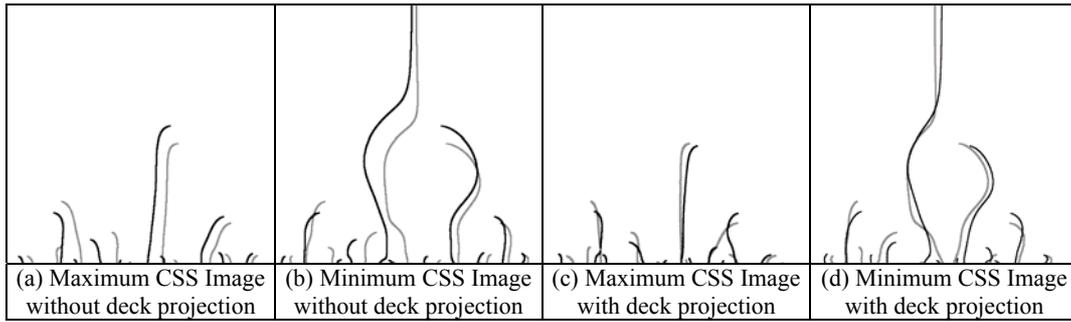

| (a) Maximum CSS Image without deck projection | (b) Minimum CSS Image without deck projection | (c) Maximum CSS Image with deck projection | (d) Minimum CSS Image with deck projection |

Fig. 6 Comparison between target (gray) and model (black) in Extrema CSS Image with and without deck projection

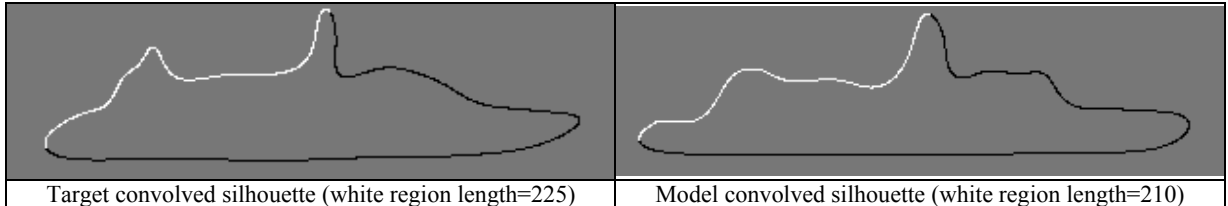

| Target convolved silhouette (white region length=225) | Model convolved silhouette (white region length=210) |

Fig. 7 Curve convolution and relative position of the selected maximum.

**2.3. Locality of the curvature measure. Improved robustness by means of lobe concavity**

Curvature is inherently a local measure. Small contour variations, especially at high smoothing degrees, could displace the curvature extreme almost freely around a section of its arc length, which results in a poor way of describing lobe characteristics. In order to overcome this problem, we have replaced the value of the curvature extrema as a means to describe a simple concavity/convexity with another parameter (C in Fig. 8) related to the curve distance to the line defined by the lobe delimiting zero crossings ($Z_0$, $Z_1$). In particular, we describe the arc by means of both the location and the distance to the zero crossing line of the arc point that lies at maximum distance of this line (see Fig. 8). These descriptors correlate better with the macroscopic appearance of the arc, and have proven to be more robust against small variations in shape. A variation of these features, in which mean instead of maximum deviation has been considered, has already been proposed in [7], as a means to filter out shallow concavities.

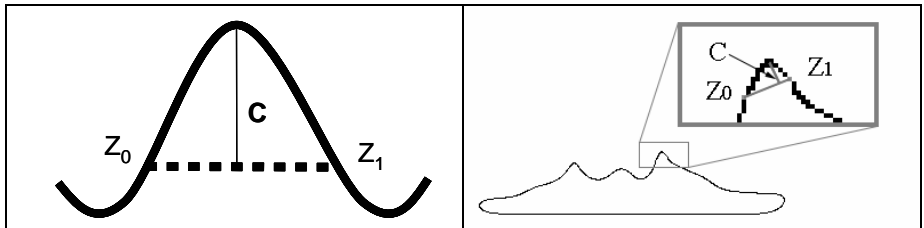

Fig. 8 Determination of lobe concavity

The improvement in the identification performance obtained when replacing the curvature by the concavity has been experimentally established, yielding a remarkable increase in the correct identification ratio.

Using concavity instead of curvature in the synthesizing process of CSS images gives place to our proposed representation: Concavity-Convexity Scale Space (CCSS) images. Curves in this representation are more stable than those in CSS ones, solving the problem of curvature extrema displacement previously mentioned. Fig. 9 visually compares CSS and CCSS images for a selected warship, remarking the structural differences between both types of representations.

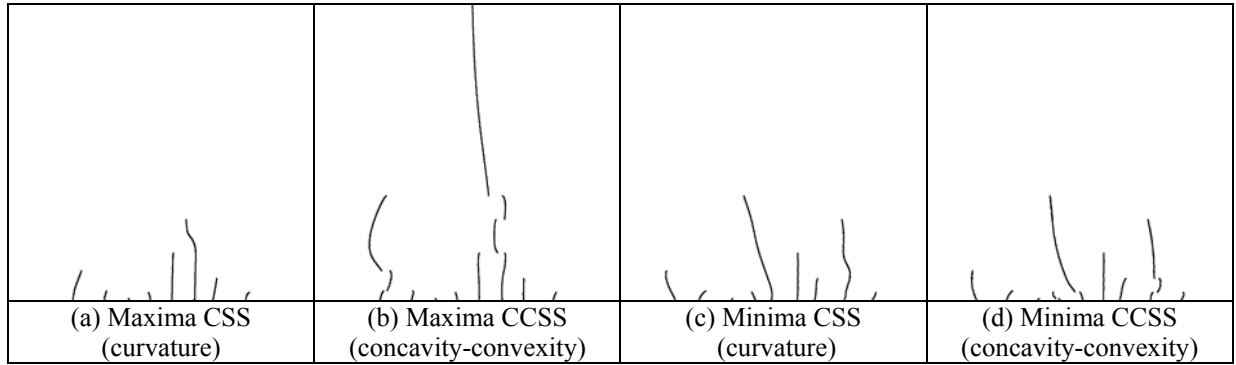

Fig. 9 Visual comparison of CSS and CCSS images

### 2.4. The effect of shallow concavities

This problem was previously analyzed in [7]. The present work proposes a new approximation in accordance with the characteristics of our own specific environment. Here, the problem appears in its most acute form in long straight regions of the warship silhouette. In Fig. 10 the silhouettes of two conflictive vessels are showed. Both silhouettes are identical, except for a minute 1-pixel step located on the straight section of the deck of the second warship.

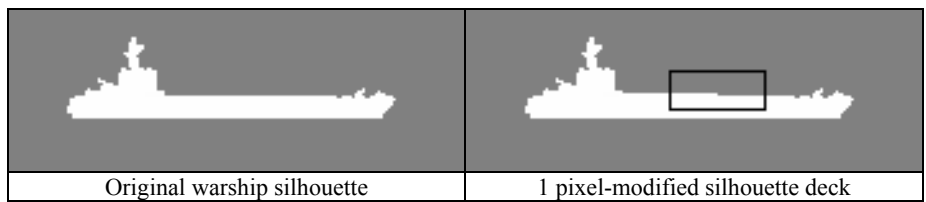

| Original warship silhouette | 1 pixel-modified silhouette deck |

Fig. 10 Comparison between two warships affected by the problem of shallow concavities

In Fig, 11, traditional CSS representations for the warships shown in Fig. 10 have been depicted. The slight variation in the deck mentioned above has as a consequence the appearance of a spurious lobe of considerable size, which obviously hinders any further attempt of comparison.

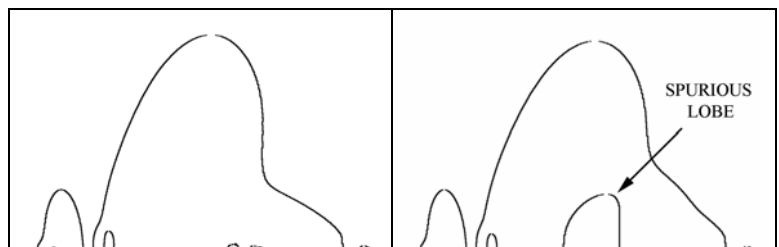

Fig. 11 Traditional CSS images for warships of Fig. 8

If extrema CCSS images are displayed for both ships, some spurious curves are also obtained, as can be observed in Fig. 12. These curves have revealed themselves as a cause of misclassification and need to be filtered out. This can be achieved considering that, to be relevant, a CCSS curve must be associated with an important enough concavity (convexity) in the corresponding warship silhouette. Accordingly, CCSS points with an associated lobe concavity below an empirically determined threshold are filtered out prior to classification. In this context it is worth to mention that the use of lobe concavity, with its enhanced stability over that of curvature values, has significantly contributed to the reliability and robustness of this filtering step.

In Fig. 13 the concavity-convexity filtered images corresponding to both ships are presented. As it can be seen, the filtering process has completely eliminated the curves corresponding to the spurious lobe, while the rest of the CCSS structure remains almost unaltered. After filtering, both images look remarkably similar.

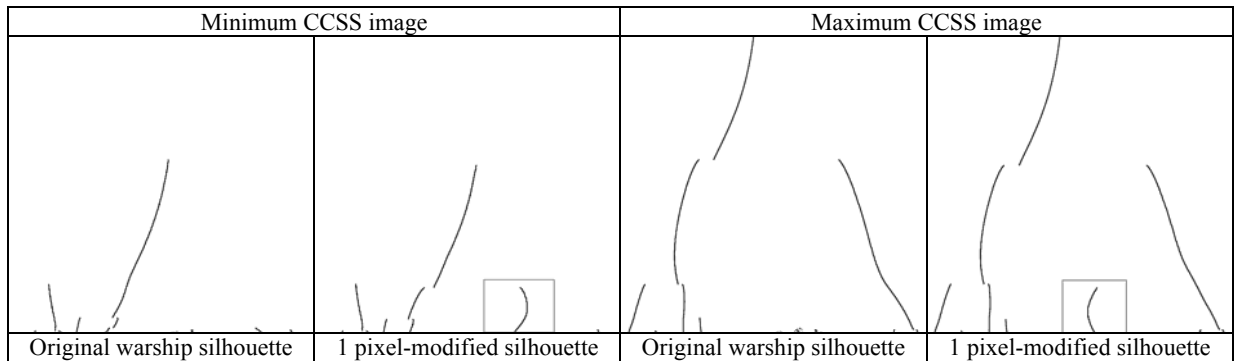
Fig. 12 Extrema CCSS images for original and 1 pixel-modified silhouettes

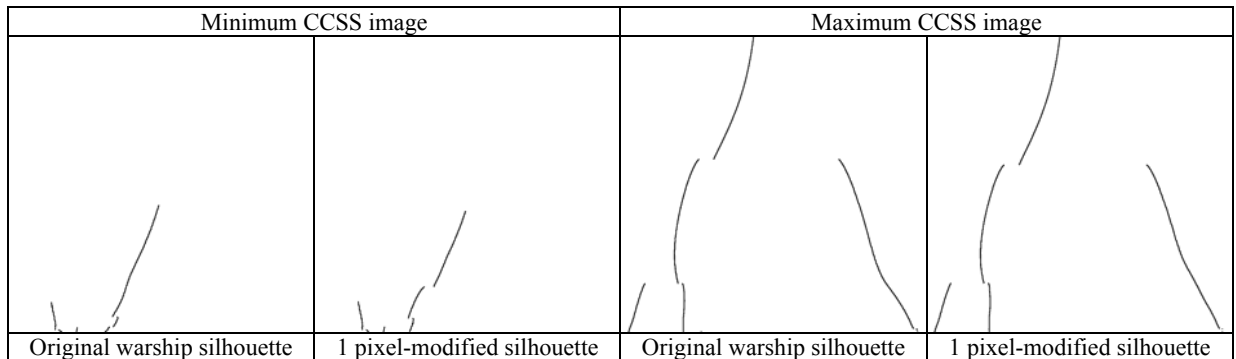
Fig. 13 Extrema CCSS images for original and 1 pixel-modified silhouettes with concavity-convexity thresholding

## 3. DESCRIPTION OF THE METHOD

### 3.1. Silhouette preprocessing

Prior to computing the concavity-convexity scale space representation (CCSS), both the target and the model silhouettes undergo a preprocessing procedure. The preprocessing of the model silhouette can be carried out off-line, previous to the identification operation.

The goal of the preprocessing stage is to increase the stability of the CCSS representation by filtering out noise-derived shape artifacts or deck objects (such as small antennas) that will not be usually distinguishable in the target at the operational observation distances and which, in any case, may often be subjected to modifications during the vessel's lifetime.

The preprocessing stage consists of the sequential application of three steps. Firstly, a morphological filtering is applied with a threefold aim: remove bright elongated objects from the deck (small antennas, noise-induced bumps, etc.), fill in small streaks on the deck and eliminate holes in the shape prior to contour extraction. The first operation is carried out by means of a morphological opening, which has been modified to avoid the creation of isolated regions. To achieve this, a morphological reconstruction method has been used, similar to that described in [8]. The second operation is performed using a classical morphological closing. The third operation is solved by means of a 4-connected background extraction followed by the determination of the complementary region. Fig. 14 summarizes the preprocessing stage by displaying the silhouettes of the raw target and model as well as the resulting ones obtained after applying the morphological filtering. As can be observed, the filtering process noticeably increases the perceptual similarity between both shapes.

Once the morphological filtering has been applied, the contour is extracted using a classical contour following algorithm with backtracking [9]. Finally, the location of the bow and the stern are determined, and the silhouette is normalized with respect to arc length.

| Original silhouette of target warship | Original silhouette of associated model |
|---|---|
| 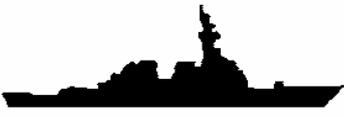 | 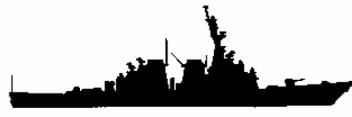 |
| Pre-processed silhouette of target warship | Pre-processed silhouette of associated model |
| 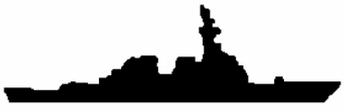 | 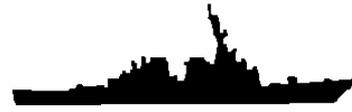 |

Fig. 14 Target and model silhouettes before and after morphological filtering

### 3.2. Concavity-Convexity Scale Space (CCSS)

Concavity-convexity curves are computed by applying a two-pass method at each smoothing scale. In the first pass, the curvature at each silhouette point is computed, and curvature zero crossings are located. In the second pass, the line defined by each pair of neighboring zero crossings is determined. This pair of points delimits a silhouette arc corresponding to a single concavity or convexity. The distance between each point of the silhouette arc and the line defined by the zero crossing pair is computed, retaining the location and the distance of the point farthest apart from the line. This distance value is affected by the curvature sign at that point, thus enabling the distinction between concavities and convexities.

### 3.3. Matching process and computation of the model-to-target assignment cost

The aim of the matching process is to obtain an optimal assignment of the CCSS points of the target image to those of a model. Under this assignment, a matching cost could be computed, numerically describing the similarity of the target CCSS curves to that of a certain model. Applying this procedure to all models in the database, a matching cost can be associated to each target-model pair, which can then be sorted in terms of similarity to the target. The models that ranked among the top ones are then presented to the operator, who makes the final decision.

The matching process is performed in two steps. In the first step, the curves corresponding to the model and the target are relatively shifted to optimize their correspondence. In the second step, all potential assignment of target and model curves are systematically explored, computing the cost of each assignment. The assignment corresponding to minimum cost is the one kept. A more detailed description of these steps is presented below.

**i) Shift correction**

Sometimes, visibility and noise perturbations could induce slight inaccuracies in the determination of bow and stern. This can cause a shift in the CCSS curves, which subsequently distorts the computation of the matching cost between target and model. Hence, this shift should be corrected before proceeding with the matching process. Obviously, this shift must be checked to be homogeneous for all curves of the target image, to avoid sweeping out real target-to-model horizontal differences in the CCSS curves.

First, maximum CCSS curves, for both the model and the target CCSS images, are analyzed following a line-by-line approach. Each point in the target CCSS image is associated to a point in the same line of the model CCSS image, where the criteria is that point-to-point horizontal distances are to be minimized. This process is repeated for each point in the CCSS target image, obtaining an average distance to model curves.

To improve the reliability of the obtained shift correction, we repeat the process for minimum CCSS images, obtaining a second distance average. These two distances are averaged and the result is applied to correct for shift between target and model curves. Application of the shift correction procedure to non-corresponding target and model curves will generally lead to large differences between the two computed distances, meaningless average shift corrections and final penalization of these incorrect associations.

**ii) Matching cost calculation**

Once the horizontal distance has been corrected, the core of the matching process can then be applied. CCSS images will be considered at this point as a set of rows of points of interest. The global matching cost will be obtained from partial point-to-point row matching cost, where each step is repeated both for maximum and for minimum CCSS images.

For each row in the target and model CCSS images, two lists of relevant maximum (minimum) points are extracted to be compared. Concavity associated values have revealed to be essential for matching cost estimation, therefore a pair (x, c) will be stored for each maximum (minimum) points in the actual row. Comparison cost is then obtained by using an exhaustive method to explore the solution tree: the Recursive Matching Matrix (RMM).

The set of relevant row points for a given image will be noted as *I*, while the set of relevant row points for the model will be referred to as *M*. Therefore, $I(i, x)$ represents the horizontal position of the $i^{th}$ extrema in the current row and $I(i, c)$ represents the concavity value of the $i^{th}$ extrema in the current row. Similar considerations apply to M.

To exhaustively search the space of potential matches, RMM must have a number of columns that exceeds or equals the number of rows. Let us denote by |I| and |M|, respectively, the cardinality of sets I and M. If |I| > |M|, RMM will be (|M| x |I|), otherwise, RMM will be (|I| x |M|). Without loss of generality, let us suppose that the first case holds (the case |M| ≥ |I| can be treated similarly). Each cell (i, j) of RMM will store the result of evaluating the following expression:

$$RMM(i, j) = [\alpha * abs(I(i,x) - M(j,x)))] + [(1-\alpha) * abs(I(i,c) - M(j,c))], \qquad (1)$$

where *abs* stands for absolute value, and α represents a scale factor between distance and concavity. A value of α=0.2 has demonstrated good empirical results.

```
Matching Cost Algorithm
INPUT: Recursive Matching Matrix (RMM)
OUTPUT: RMM Associated cost
Begin

        OPTCost = ∞
        TCost = 0

        If RMM has no rows, return 0.

        For each column in the RMM, ci
                Take first row in RMM, r1
                Construct RMM2 by eliminating r1 and ci from RMM
                TCost = RMM(r1, ci) + Call recursively with RMM2

                If TCost < OPTCost update OPTCost to TCost
        End for

        Return OPTCost

End
```

Fig. 15 Pseudocode for Recursive Matching Matrix Cost Calculation

The algorithm in Fig. 15 provides optimum cost-for-row matching. It is still needed to add a penalty term to the cost to take into account those curves in the target that have no correspondence in the model and vice versa:

$$OPTCost_f = OPTCost + ( abs( M - N ) * \sigma ), \qquad (2)$$

where *OPTCost* is the cost obtained by application of algorithm described in Fig. 15, and σ is a gain factor. Empirically, good results have been obtained with a value of σ = (70 * α), being α the scale factor between distance and concavity previously mentioned.

By repeating this execution procedure for each row in actual CCSS images and then adding up the resulting costs, the total matching cost will be obtained for the maximum-CCSS image. Minimum-CCSS image matching cost can be obtained in a similar way. By adding maximum and minimum partial costs, a global target-model cost is obtained.

### 3.4. Solution set sorting

Once obtained the global target-to-model cost for each model in the search space, the models are sorted in terms of their similarity to the target to be identified. The models ranked first in this sorted list are displayed to the operator, who makes the final identification decision.

## 4. EXPERIMENTAL RESULTS

The proposed method has been applied using a search database composed of 1129 vessel silhouettes generated from scanned line drawings of world warships [10]. The developed decision support system prototype allows the user to extract a vessel silhouette from a real image and present it as input data to the search engine. The output of the system is the 1129 silhouettes set, sorted in terms of the probability of matching the original image.

Table 1: Frequencies table showing system results for 50 warships test set.

|  | 1st pos. | 2nd pos. | 3rd pos. | 4th pos. | 5th pos. | 6th pos. | Other |
|---|---|---|---|---|---|---|---|
| Relative Frequency | 30 | 9 | 3 | 4 | 3 | 1 | 0 |
| Relative Cumulative Frequency | 30 | 39 | 42 | 46 | 49 | 50 | 50 |
| Absolute Frequency | 0.6 | 0.18 | 0.06 | 0.08 | 0.06 | 0.02 | 0.0 |
| Absolute Cumulative Frequency | 0.6 | 0.78 | 0.84 | 0.92 | 0.98 | 1.0 | 1.0 |

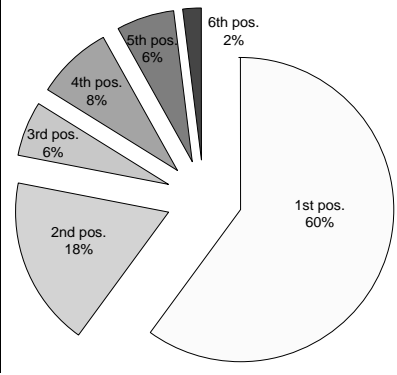

A set of 50 warship images acquired in real operational conditions has been used to test the system. The silhouettes corresponding to each of these images were extracted and injected to the system, which sorted the 1129 vessel database in terms of similarity to the input silhouette. Table 1 lists the frequencies at which the matching model appears in a given position of the sorted database. This information is graphically displayed on the adjacent figure. As it can be seen, the correct warship models appear with a probability close to 80% in the first two positions of the sorted database. All input silhouettes were correctly identified in the first six positions of the database. Finally, Fig. 16 shows four examples of system warship silhouette identification and the four-most similar results for each one of the warships. Each column corresponds to a different target image, where the top rows show the original image and the extracted silhouette, while the bottom rows display the four silhouettes of the models that are most likely to match the original vessel. The correct warships appear highlighted in the 1st, 1st, 2nd and 3rd row respectively form left to right. In the last two cases, the identification could be considered to yield optimum results, as only models corresponding to the same class are ranked above the matching one. This fact can be visually checked in Fig. 16.

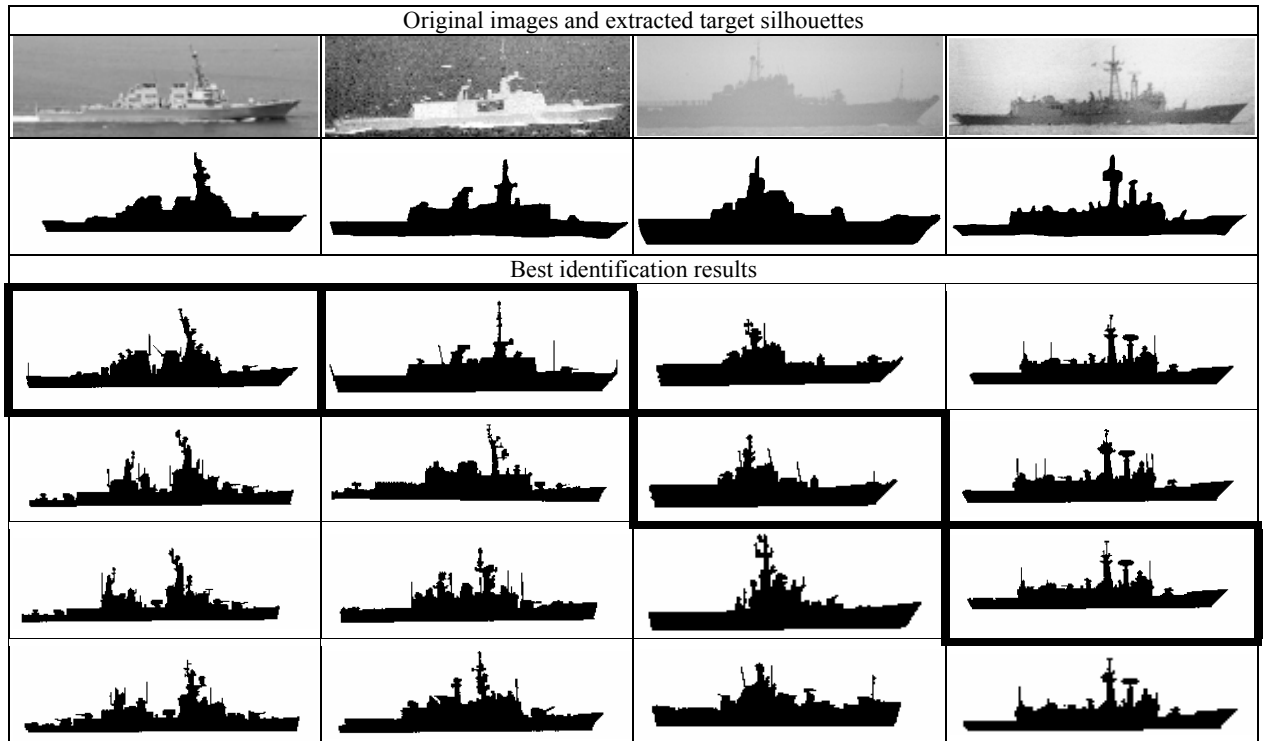

Fig. 16 Original images, extracted silhouettes and top-ranked results for warships from the test set

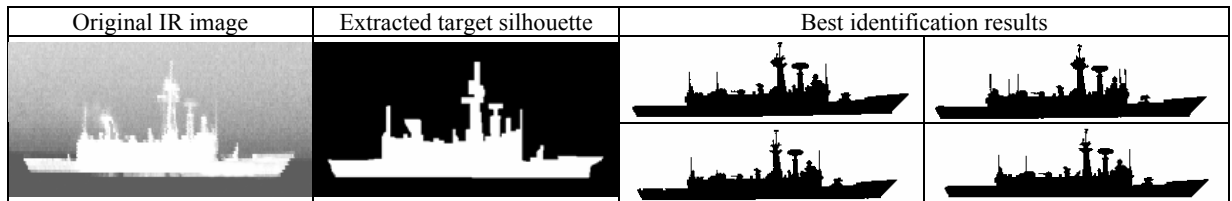

Fig. 17 Warship identification example for infrared image

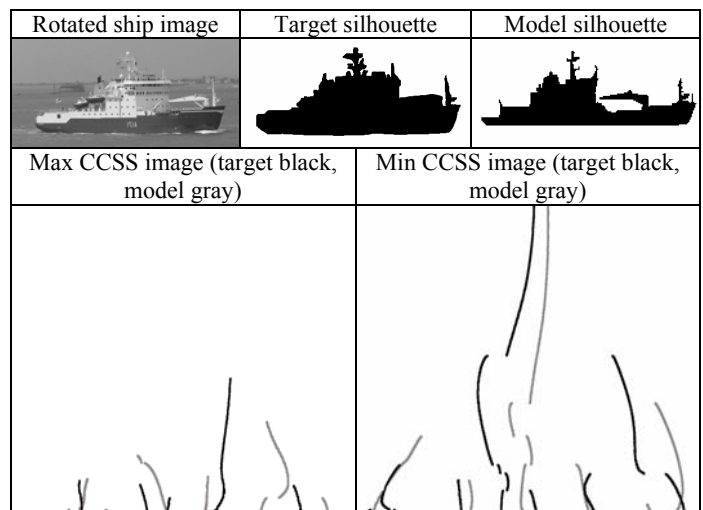

Fig. 18 Viewpoint angle effects in CCSS method

The type of sensor used in warships image capture has empirically showed to have little relevance as long as the silhouette is visually recognizable. Test cases included some infrared (IR) and intensified images (II), with which results of similar quality were achieved. Fig. 17 shows an IR test case and his respective four best results. The correct warship appears in the 2$^{nd}$ place in the 1129-warship sorted output set. The first result corresponds to a ship of the same class.

One of the most important inheritances from the curvature scale space method is that of noise tolerance. This feature is essential in silhouette identification. Different system operators or silhouette extraction algorithms will extract different silhouettes, what theoretically could be modelled as a noise factor in the respective ideal warship image boundary. The quality of the results obtained in this direction supports the goodness of the approximation.

This method has revealed to be affected by large modifications in the viewpoint angle. Best performance is obtained with an image capture viewpoint angle near 90º with respect to ship direction. Anyway, the system has demonstrated an optimum behaviour for angle variations within ± 10º with regard to profile view. Fig. 18 shows the differences between the rotated and the ideal extracted silhouette, as well their respective CCSS images. This rotation, which is larger than 10º, causes in this case the misidentification of the warship.

## 5. CONCLUSIONS

A new system to support identification of military vessels has been reported. The system is based on an evolution of the Curvature Scale Space (CSS) method, which has lead to a significant increase in the performance and robustness of the system. This novel approach, the Concavity-Convexity Scale Space (CCSS) representation, overcomes problems such as lobe-splitting, model-to-target silhouette arc length variations, shallow concavities and curvature artifacts that preclude the direct use of the CSS representation to solve this problem.

This CCSS representation has been embedded on a complete identification system comprising the steps of silhouette preprocessing, computation of the CCSS representation, matching and model database sorting. The system has been applied on operational imagery acquired with sensors operating in different spectral ranges, proving the high performance and robustness of the developed method.

## 6. ACKNOWLEDGEMENTS

The authors gratefully acknowledge the financial support provided by the Subdirección General de Tecnología y Centros (SDGTECEN) of the Spanish Ministry of Defence.